\documentclass[conference]{IEEEtran}
\usepackage{times}
\usepackage{todonotes}
\usepackage{ulem}
\usepackage{multirow}
\usepackage{balance} 
\usepackage{booktabs}
\usepackage{colortbl}
\usepackage[numbers]{natbib}
\usepackage{multicol}
\usepackage[bookmarks=true]{hyperref}

\begin{document}

\title{Unsupervised, Bottom-up Category Discovery for Symbol Grounding with a Curious Robot}

% You will get a Paper-ID when submitting a pdf file to the conference system
\author{\authorblockN{Catherine Henry}
\authorblockA{Department of Computer Science\\
Boise State University\\
Boise Idaho, 83725\\
catherinehenry@u.boisestate.edu}
\and
\authorblockN{Casey Kennington}
\authorblockA{Department of Computer Science\\
Boise State University\\
Boise Idaho, 83725\\
caseykennington@boisestate.edu}
}

% avoiding spaces at the end of the author lines is not a problem with
% conference papers because we don't use \thanks or \IEEEmembership

%Homer Simpson\authorrefmark{2},
%Montgomery Scott\authorrefmark{3} and
%Eldon Tyrell\authorrefmark{4}}
%Telephone: (800) 555--1212, Fax: (888) 555--1212}
%\authorblockA{\authorrefmark{4}Tyrell Inc., 123 Replicant Street, Los Angeles, California 90210--4321}}

\maketitle

\begin{abstract}
Towards addressing the Symbol Grounding Problem and motivated by early childhood language development, we leverage a robot which has been equipped with an approximate model of curiosity with particular focus on bottom-up building of unsupervised categories grounded in the physical world. That is, rather than starting with a top-down symbol (e.g., a word referring to an object) and providing meaning through the application of predetermined samples, the robot autonomously and gradually breaks up its exploration space into a series of increasingly specific unlabeled categories at which point an external expert may optionally provide a symbol association. We extend prior work by using a robot that can observe the visual world, introducing a higher dimensional sensory space, and using a more generalizable method of category building. Our experiments show that the robot learns categories based on actions and what it visually observes, and that those categories can be symbolically grounded into. 
\end{abstract}

\IEEEpeerreviewmaketitle

\section{Introduction}
The \textit{symbol grounding problem} posits that the meaning behind all words cannot be entirely composed of other words (e.g., definitions); at some point something else has to give meaning to a sufficient subset of words that are \textit{grounded} into the physical world \cite{Harnad1990-fr,Vincent-Lamarre2016-gi}. The meaning of such foundational words is often derived from physical experience with concrete objects within the physical world in a bottom-up conceptual learning progression \cite{Muraki2022-nw}. In other words, the learning of conceptual categories comes first and the symbol (i.e., word label) comes later, whereas in most research on symbol grounding, the symbol comes first. This unsupervised, bottom-up approach, where a category develops independent of the symbol which is then later attached to the category, follows the early language learning progression of human children. According to \citet{Smith2005-qt}, children live in a physical world full of rich regularities that organize perception, action, and thought and they interact in a social world to learn a shared linguistic, symbolic communicative system. This means that physical embodiment is very likely a prerequisite for symbol grounding (i.e., learning categories bottom-up) and, therefore, language acquisition (see also \cite{Harnad2017-jo,Bruner1983-uv,McCune2008-oy,Eimas1987-ef,Cangelosi2015-jl}). We infer, therefore, that the most likely platform for language acquisition that matches closely to humans will be on robotic platforms because they are embodied and operate in the same world that humans do.\footnote{This is in contrast to how language models learn using a distributional representation of word meaning largely from text or text-image pairs; the learning progression is the opposite that of humans \cite{Schlangen2023-sc}.} 

But what motivates children to explore and categorize their world? In this paper, we focus on \textit{curiosity}. As \citet{Ginsburg_and_Sylvia_Opper1988-uo} explained, ``curiosity is a function of the relation between the new object and the individual's previous experience." As a child explores and engages with the space around them, they are likely to encounter objects which may be entirely novel or which they have not yet interacted with while a situational expert (i.e., adult) is present to assign a symbol (i.e., a word) to the object of interest--following the symbol grounding progression. For example, a child picks up an un-before-seen object to see and touch it, but does not learn until later when an adult utters \textit{spoon} that there is a label for that category. 
% In \cite{Piaget1952-kp} Piaget makes the general appraisal of a child's spontaneous visual interests ``...the subject looks neither at what is too familiar, because he is in a way surfeited with it, nor at what is too new because this does not correspond to anything in his schemata (for instance, objects too remote for there yet to be accommodation, too small or too large to be analyzed, etc). In short, looking in general and the different types of visual accommodation in particular are put to use progressively in increasingly varied situations. It is in this sense that the assimilation of objects to visual activity is 'generalizing'". 

% The process of labeling categories is required for learning language: imagine a child observing cows in a field during a daily walk in a stroller. Without a situational expert actively engaged and holding joint attention with the child to assign a label, the visual experience with an object might result in a loose conglomeration of one or more senses into a vague category with no direct symbol association, something which we consider to be an ``unlabeled category". 

In this paper, we leverage an existing model of curiosity from Oudeyer et al \cite{Oudeyer2007-av,Oudeyer2006-bj,Oudeyer2005-yi} and a Cozmo robot which can physically move through a space and explore the visual properties of objects contained within that space in order to build unsupervised bottom-up categories. The physical movement of the robot, per action/perception turn, consists of a single rotation and linear movement forward or backward. These movements are an approximation of an infant's motor actions of \textit{looking} and \textit{crawling}. Exemplars (realized as vectors) collected throughout the exploration process result in a representation of the sensorimotor space $\mathbf{SM(t)}$ are composed of the motor action and the sensory experience of the robot associated with each movement. The sensory experience of the robot is a visual vector and can also include supplementary positional information. As the robot explores, there is a process in which categories are carved out of the sensorimotor space. The robot's autonomous, unsupervised discovery and categorization of these \textit{regions} of sensorimotor space is the primary focus of our work as it matches more directly with the bottom-up learning progression of children and, we argue, the symbol grounding problem of identifying categories before they are labeled. 

With this \textit{curious} system, it is our intent to flip traditional methods of grounding in that we begin with an existing, unlabeled, category which the robot later associates with a symbol rather than starting with a symbol and associating a relevant category (which constitutes most work on symbol grounding). It is our hope that by enabling the robot to autonomously identify unsupervised categories throughout the space without proactively or immediately assigning labels, we are able to explore meaning with a bottom-up perspective. Furthermore, this independent exploration of a space allows for the reduced manual effort---perhaps even removal---of defining and building tasks of varying complexity for the purpose of aggregating data to be used in training systems. Our experiments build directly on and systematically extend prior work until we test our hypothesis that the categories are developed enough for symbol grounding by learning word-level classifiers for the robot's discovered categories.

\section{Related Work}

Our work is partially related to a growing literature on few-shot learning in robotic scenarios, including datasets \cite{Lomonaco2017-hh} and methods. Early methods included \cite{Xu2007-kq,Squire2007-ql} which used different Bayesian approaches for word learning. Seminal work by \cite{Lake2015-km} proposed a method to learn novel categories, focusing on character recognition using few examples. More recently, \cite{Ayub_undated-jf} proposed a Gaussian Mixture Model method for few-shot learning to mitigate catastrophic forgetting, and \cite{Nie2023-gy} proposed online active control for lifelong learning of object categories based on visual information. Finally, \cite{McClurg_undated-uz} used incremental learning and active class selection to improve few-shot learning. Though related to this literature, our research is fundamentally different in that we are directly focusing on resolving an aspect of the symbol grounding problem by bottom-up learning of unlabeled categories discovered as the robot curiously explores the world. This brings to bear an \textit{enactive} element of learning in that the robot is actively exploring its physical space instead of only being exposed to pre-collected data \cite{Johnson2008-zd,Bisk2020-wy}. In contrast, this related work is focused on categories with top-down labels. 

More directly related to our work is \textit{Intelligent Adaptive Curiosity} (IAC), which has been extended in a variety of ways. Aly et al \cite{Aly2017-bm} leverage one such extension, RL-IAC \cite{Craye2016-tl}, in that they model segmenting objects in an unsupervised manner. Our work shares additional parallels with theirs as they explore grounding spatial concepts and object categories through visual perception. 

Our work is also related to developmental robotics (see \cite{Cangelosi2015-jl}) in that we use similar methods to enable a robot to explore and build a degree of understanding about its environment, but we deviate from the developmental robotics agenda in that we focus on using these methods for ultimately building concrete categories that form the basis of first language acquisition. Thus our work is related to the long and still growing literature on addressing Harnad's Symbol Grounding Problem. Recent work showed that abstract categories can be built on more basic, concrete categories, but the categories were pre-determined \cite{Kennington2022-db}, whereas in our work categories are discovered autonomously by the robot as it explores. 

% A benefit of exploring the goals in the task space rather than the motor configurations in a motor space is that it should allow for more efficient exploration in redundant sensorimotor spaces. However, we are looking to explore the space in a redundant way as at this juncture we are far more interested in capturing a rich sensorimotor representation than the achievement of tasks. As such, in our work we have opted to deviate from continued iterations by Oudeyer and team, such as Self-Adaptive Goal Generation Robust Intelligent Adaptive Curiosity (SAGG-RIAC) introduced in 2013.

\section{Method}

In this section, we give a background of the work on curiosity that we build on, then explain our system that uses curiosity to explore and learn about its space. 
 
\subsection{Background} 
\begin{figure}
\centering
\includegraphics[scale=0.25]{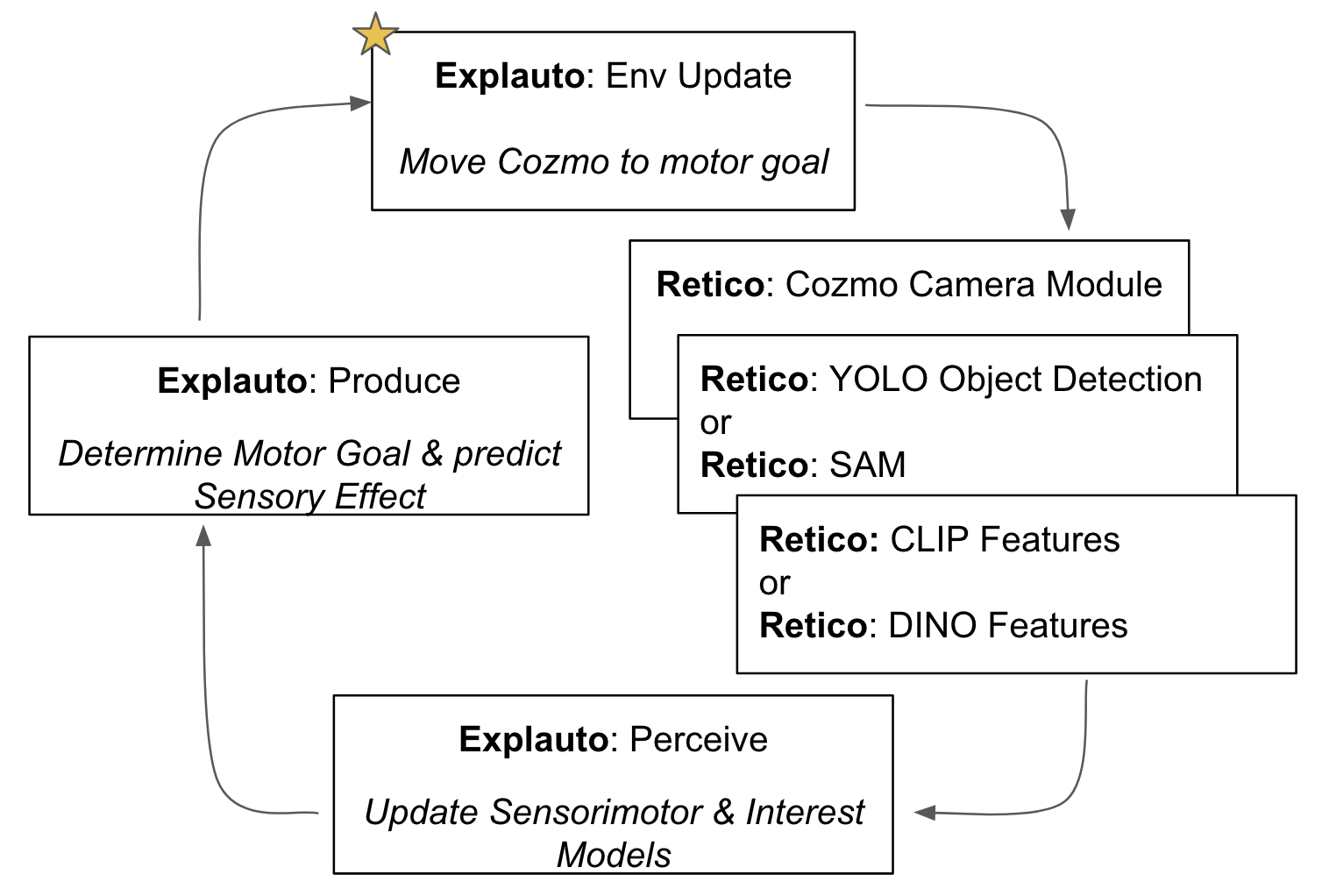}
\caption{High level overview of the Explauto + Retico pipeline. Initialization point is denoted by the star. In the Perception phase, the Sensorimotor model is updated with the Motor Goal vs Actual Sensory Effect and the Interest Model is updated with the Motor Goal + Predicted Sensory Effect vs Motor Goal + Actual Sensory Effect. Production phase uses the InterestModel to for the Motor Goal and the Sensorimotor Model to predict Sensory Effect.}
\label{fig:pipeline_overview}
\end{figure}

The learning pipeline in our system includes technologies which are expanded on in the following subsections. Figure \ref{fig:pipeline_overview} is a high level system overview referencing each of these components in our model.

\paragraph{Cozmo}
The robot we use in our experiments is Cozmo.\footnote{Originally produced by Anki, Cozmo is now Licensed by Digital Dream Labs.} Cozmo is a small tabletop robot which moves using a tread system and has a built-in camera capable of streaming at 320x240 (QVGA) resolution at a rate of approximately 15 frames per second. We leverage the Anki provided Python SDK for interacting directly with Cozmo for sensory input and robot control. 
% For the purpose of our experiments the SDK allows us to capture images from the camera feed and the control of motor actions for navigation. 

\paragraph{Retico}
We use the incremental framework Retico and the robot-ready spoken dialogue system extensions \cite{Michael2019-dx,Kennington2020-qx} to support the integration of Cozmo with our system. While Retico is typically oriented to Spoken Dialogue Systems, low granularity transfer of information (e.g., camera frames) allows for a straightforward execution of and data pass-through to and from Explauto functions. Furthermore, we are able to leverage pre-existing robot-ready spoken dialogue system modules including object detection, and object representation. Importantly, leveraging Retico's framework for this work will enable integration of capabilities that will allow the robot to learn object category names (i.e., words) from a human interlocutor, which we leave for future work.

\begin{figure} % # TODO: iirc one of the concerns from the TAHRI reviewers was too small of text on figures. What is the best way to approach that? Rework the figures or do we have space to scale them up?
\centering
\includegraphics[scale=0.2]{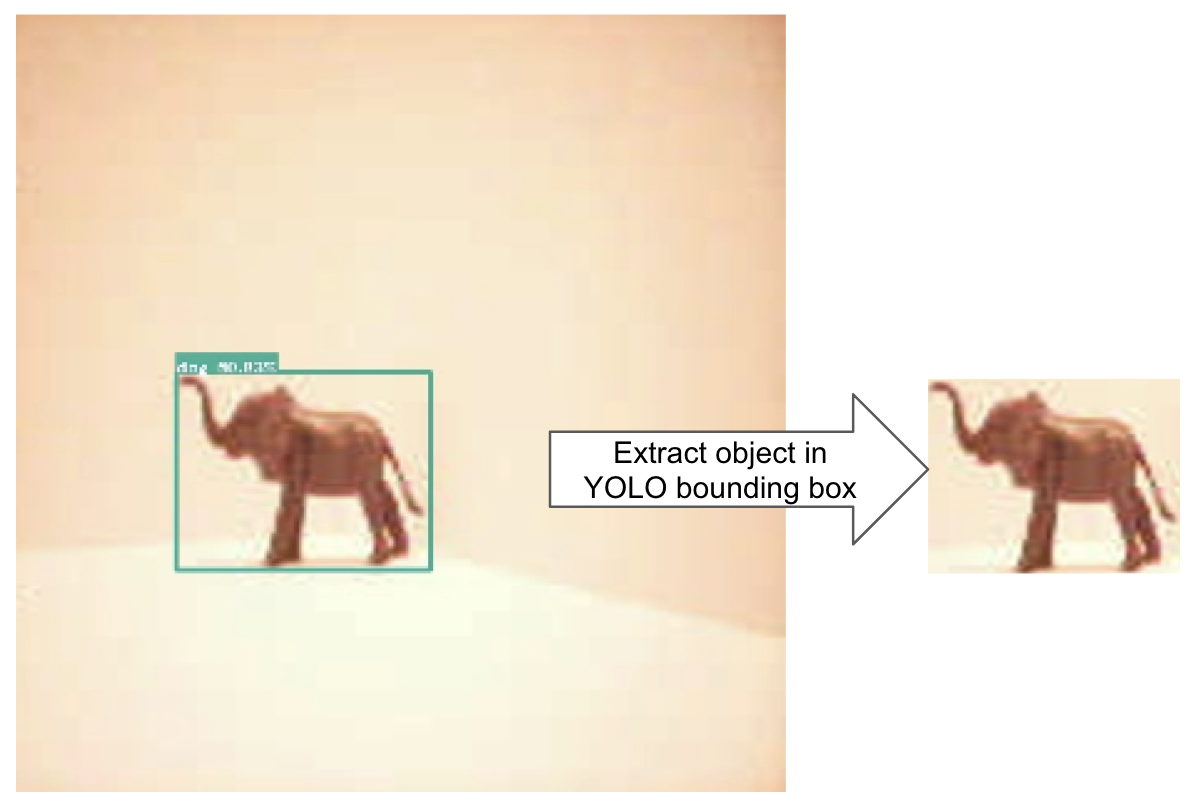}
\caption{Using the YOLO bounding box to extract a sub-image of the object in view. This sub-image is passed to the CLIP module. The object name assigned by YOLO is dropped entirely.}
\label{fig:yolo_obj}
\end{figure}

\paragraph{Object Detection \& Representation (v1): YOLOv4 \& CLIP}
YOLO (You only look once) is an object detection model designed for real-time object detection tasks \cite{Bochkovskiy2020-po}. YOLO deviates other from other detection systems by re-framing object detection as a single regression problem going straight from image pixels to bounding box coordinates and class label probabilities. YOLO allows us to quickly identify objects within an image and select observed objects with the highest confidence score, in our case the robot only observes a single object at a time in a given frame.\footnote{We are using the YOLOv4 implementation from \url{https://github.com/yuto3o/yolox} with pretrained weights from \url{https://github.com/AlexeyAB/darknet/releases}} We only use YOLO to identify that an object exists and its bounding box; we ignore its label. YOLO is meant to act as a proxy for a child's capability to isolate individual objects, even if their categories are unknown. Figure \ref{fig:yolo_obj} is an example of how we use YOLO to extract a sub-image with the object from a camera frame. 

CLIP is a multi-modal vision and language model that can be used for image-text similarity and for zero-shot image classification \cite{Radford2021-pk}. We leverage the image encoder to encode the sub-image from the YOLO bounding box with a resulting feature vector of size 512. This vector, supplemented with additional positional information including width, height, and object location relative to the overall image, serves as the backbone representation of our unlabeled categories. We refer to this vector as CLIP+.

YOLO is trained using supervised data with few object categories and CLIP is a model that is highly trained on symbolic language data, meaning they are somewhat at odds with our long-term research goals that adhere to child language acquisition settings. YOLO and CLIP act as placeholders in our first experiments to allow us us to systematically ensure that our pipeline has the highest likelihood of success as we examine systematic changes from prior work. For our final experiment, we use models for object detection and representation that are modeled and trained using unsupervised methods, explained presently. 

\begin{figure}
\centering
\includegraphics[scale=0.2]{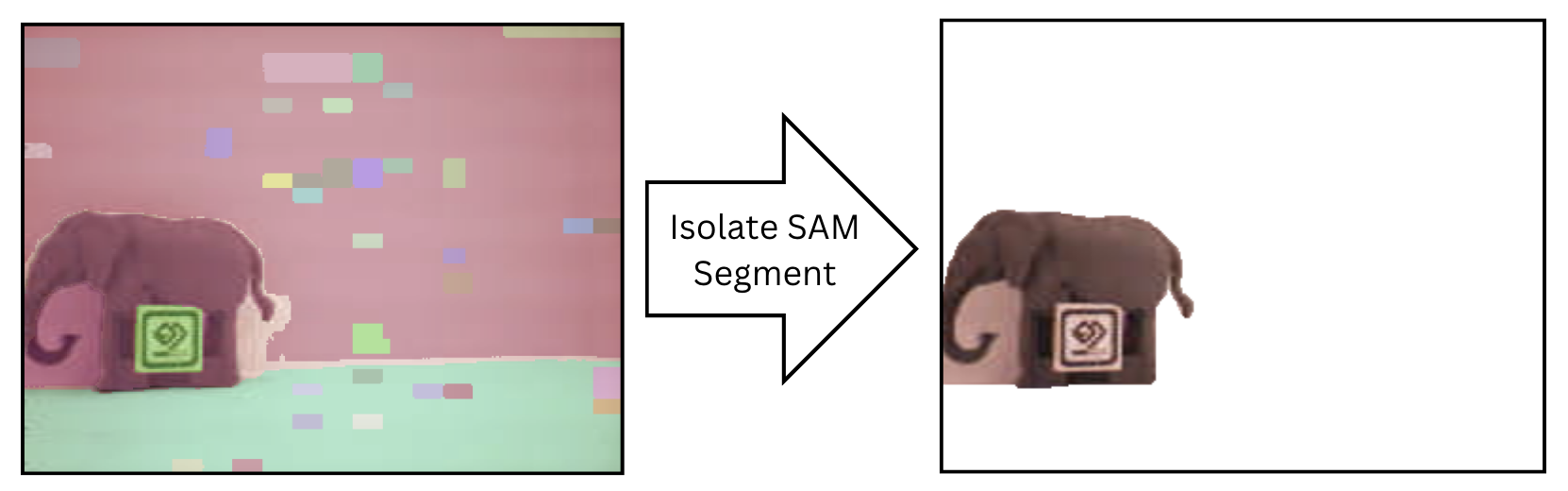}
\caption{Using the SAM segments to mask a sub-image of the object in view. This masked image is passed to the DINOv2 module.}
\label{fig:sam_obj}
\end{figure}

\paragraph{Object Detection \& Representation (v2): Segment Anything \& DINOv2}
The \textit{segment anything model} (SAM) is an image segmentation model for object segmentation tasks \cite{Kirillov_undated-wb}. SAM is not like other methods (e.g., YOLO) which identify objects with labels; rather, SAM allows us to identify objects within an image and, in our case, only observe a single ``salient" object at a time in a given frame from the robot's camera. SAM is meant to act as a proxy for a child's capability to identify unlabeled objects. Figure \ref{fig:sam_obj} is an example of how we use SAM to isolate the object through masking.

DINOv2 is a vision model we use to represent objects segmented from SAM \cite{Oquab2023-wz}. DINOv2 is a good choice for this step because the model training is largely unsupervised, similar to how children learn about the visual world more directly than CLIP.  We leverage the DINOv2 small model to encode the sub-image from the SAM segment with a resulting feature vector of size 384. This vector serves as the backbone representation of our unlabeled categories.

\subsection{Modeling Curiosity: Explauto}
The Explauto framework is the driving force behind the curiosity in our system. It is a framework for studying, modeling, and simulating curiosity-driven learning and exploration in both simulated and physical robotic systems, developed by the Inria FLOWERS research team \cite{Baranes2009-jn,Moulin-Frier2014-gx,Oudeyer2005-yi}. Explauto comes with several sensorimotor systems available including simulated environments and interfaces to real robots.\footnote{Based on Dynamixel actuators through the Pypot library} The Cozmo robot is not natively integrated with Explauto, for example it does not have positional information of an arm end-effector. While we do retrieve the camera output directly from Cozmo, which we task Explauto to use, the sensory data we use in our model requires  postprocessing of the camera output through object detection and representation. 

% Rather than building out support for Cozmo, SAM, and DINOv2 in the Explauto framework and leveraging the provided Experiment functionality, we opted to do a majority of our modeling and development within the Retico system which already had modules available supporting the Cozmo SDK, SAM, and DINOv2. This resulted in only minor adjustments to the Explauto framework for our base system to be functional. 

% \begin{figure}
% \includegraphics[scale=0.3]{good_yolo_group.png}
% \caption{A collection of examples where the YOLO bounding box captures the expected portion of the object in view.}
% \label{fig:yolo_obj_good}
% \end{figure}

The Explauto architecture consists of three core processing levels which we elaborate on in the System Overview below. While the Explauto framework was designed to support both motor and goal `babbling' as well as inverse and forward predictions, our implementation focuses only on motor babbling with forward predictions. As such, in the following sections we remain focused on the functionality relevant to our experiments, though it is important to note that this only captures one aspect of the overall Explauto system.

\begin{figure*}[h]
\centering
\includegraphics[scale=0.35]{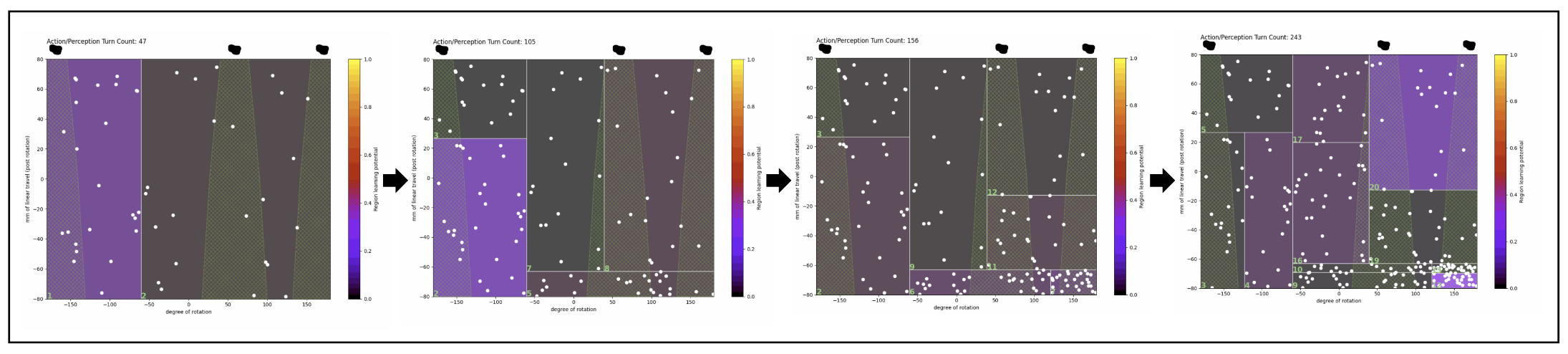}
\caption{With each iteration the Learning Potential for the region updates dynamically. Regions with high learning potential have brighter colors. 
% Because we use the Learning Progress Maximization sampling procedure with the epsilon\_greedy mode, regions with high learning potential are likely to be sampled from excluding a small espilon likelihood of random sampling.
}
\label{fig:learning_potential_progression}
\end{figure*}

The \textbf{Sensorimotor Model} learns mappings between the robot motor actions and the sensory effect they produce. This is an active, unsupervised model which is trained iteratively on the sensorimotor experience. This model is also used in performing the forward predictions given a motor command. The sensorimotor model maintains a dataset which includes a list of motor commands and sensory effects from each execution as well as a KD Tree of the motor commands.\footnote{The KD Tree implementation is from \url{https://docs.scipy.org/doc/scipy/reference/generated/scipy.spatial.cKDTree.html}} Explauto provides several sensorimotor models including locally weighted linear regression (LWLR), which we use in our implementation. The LWLR sensorimotor model predicts the sensory effect $s$ given a motor command $m$ by computing a linear regression of \textit{k} nearest neighbors of \textit{m}. The KD Tree supports a nearest neighbor lookup, which allows for indexing the motor and sensory lists to extract the \textit{X}, \textit{Y} values used in this linear regression. 

The \textbf{Interest Space} is the area of exploration from which goals are sampled; for our purposes we are using motor babbling strategies so the interest space consists of four possible motor actions $M$: \textit{degrees of rotation} (left or right) and \textit{mm linear travel} (forward or backward).

\begin{figure*}[h]
\centering
\begin{minipage}[c]{0.45\linewidth}
\centering
\includegraphics[scale=0.2]{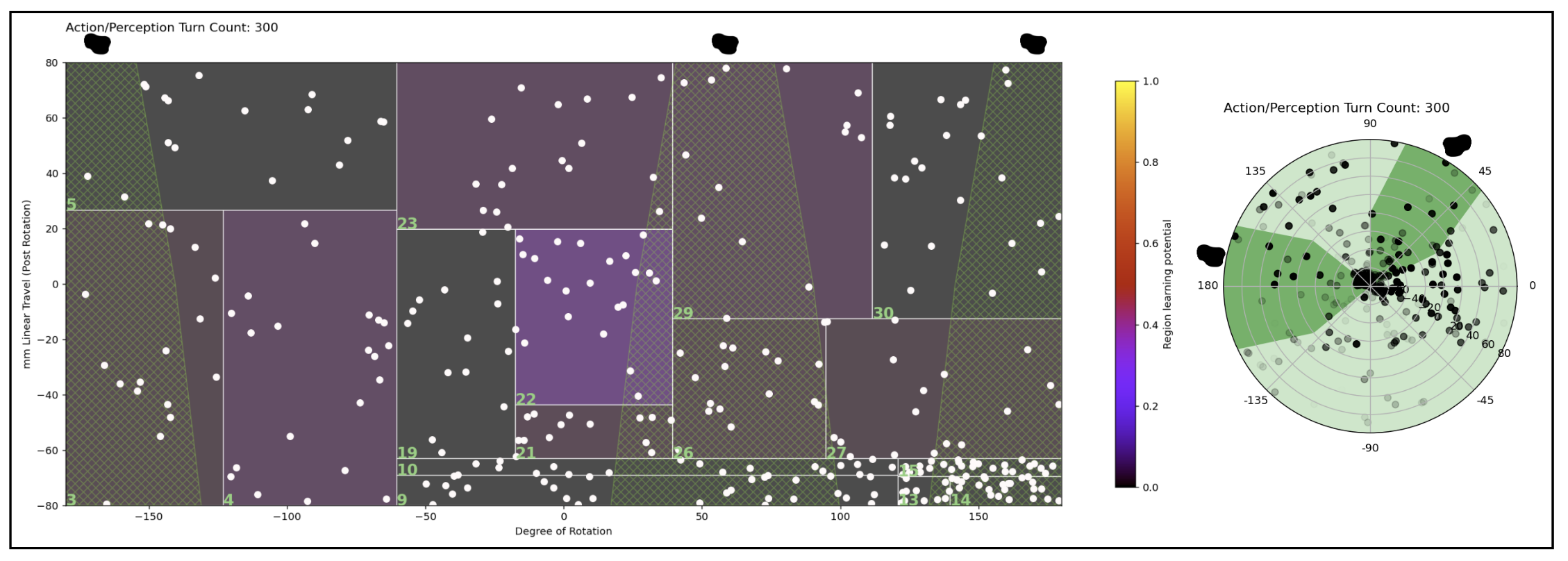}
\caption{Results from an execution of our baseline setup. In the grid plot on the left, the sub-regions (unlabeled categories) have a very loose alignment with the objects in the space. The polar plot on the right shows that recent motor goals continue to be sampled in both object and non-object areas with a heavier concentration of activity at the a negative mm Linear Distance. }
\label{fig:experiment_a_plots}
\end{minipage}
\hspace{0.3cm}
\begin{minipage}[c]{0.45\linewidth}
\centering
\includegraphics[scale=0.2]{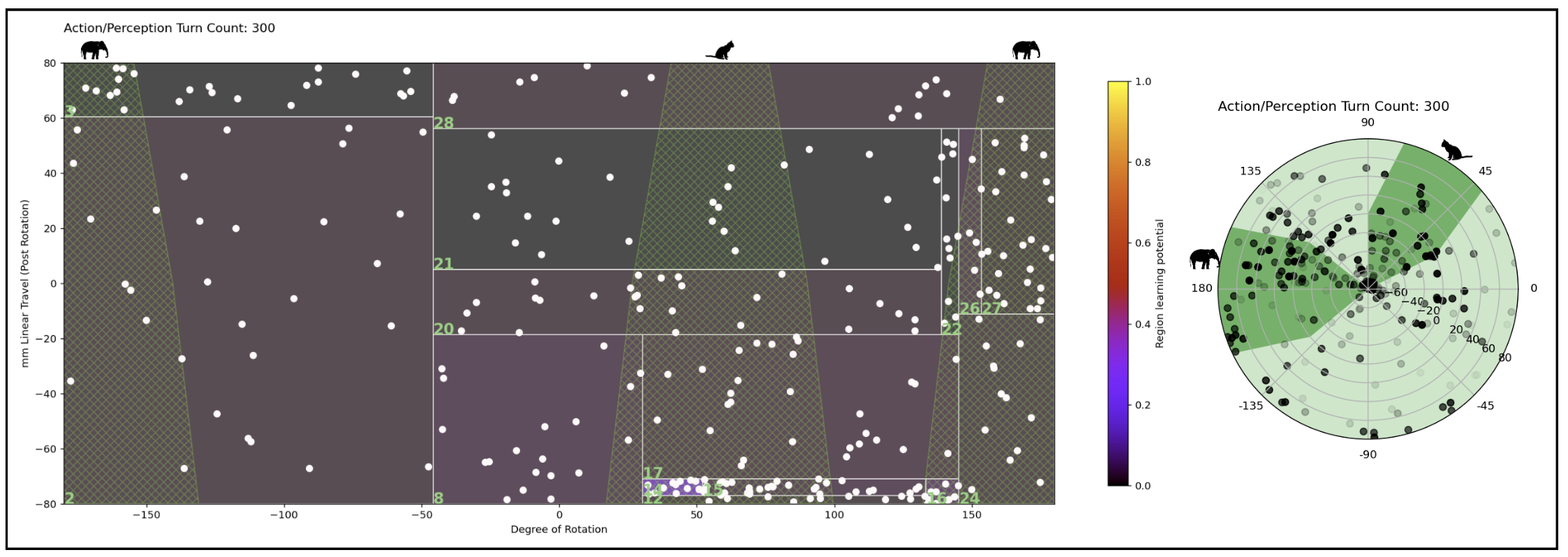}
\caption{Results from an execution of Experiment 1. The grid plot to the left shows the sub-regions continue to loosely align with the objects in the space and with fewer sub-regions developing in non-object areas than the Baseline in 
Figure \ref{fig:experiment_a_plots}. The polar plot to the right shows recent motor goals are more frequently around the objects with fewer goals being sampled in the large whitespace zone. }
\label{fig:experiment_b_plots}
\end{minipage}
\end{figure*}

% \paragraph{Interest Model: Splitting Categories}
The \textbf{Interest Model} implements the active exploration process and it is what approximates curiosity in our learning system. It is responsible for selecting motor goals from the Interest Space which improve the predictions of the Sensorimotor model. The sensory results associated with the execution of the motor goal are added to the experiential history stored in the Sensorimotor Model. At each time step, the sampling procedure selects a motor action and the sensorimotor model attempts to predict the sensory outcome. A `measure of competence' is calculated using the actual and predicted sensory outcomes and the competencies for a region are used to calculate a region's learning progress. In our experiments, we begin with the Euclidean distance calculation then introduce a cosine distance calculation as a part of our contribution. 

\begin{figure}[h]
\centering
\includegraphics[scale=0.3]{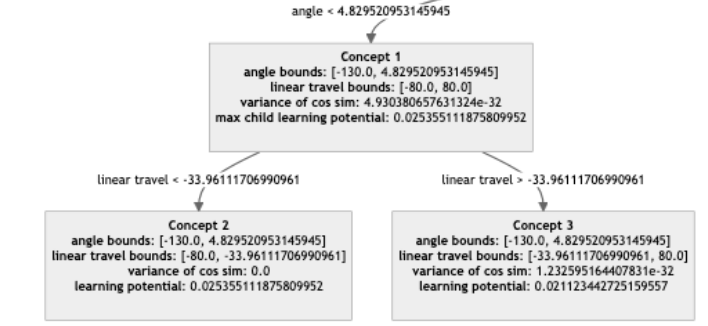}
\caption{ KD sub-tree example; the top node splits using an angle threshold whereas the bottom two nodes split on linear travel; each records information about angle and linear travel bounds, variance of the vectors (euclidean or cosine distance), and learning potential.}
\label{fig:kd_tree_example}
\end{figure}

The sampling procedure we use for exploring the interest space is \textit{Learning Progress Maximization}. This sampling procedure discretizes the interest space using a KD Tree like structure,\footnote{adapted from scipy.spatial.kdtree} where each split is along alternating axes. A region is split when the number of observations exceeds the configured maximum per region. There are various methods provided for determining the value along which to split the region, in our experiments we use the provided \textit{Best Interest} split introduced in \citet{Baranes2013-nv} as well as a \textit{Variance of Cosine Similarity} split we introduce below. A visual representation of the discretized interest space as the result of region splitting can be seen in Figure \ref{fig:experiment_c_plots} from the Experiment section below; a corresponding KD-subtree can be seen in Figure~\ref{fig:kd_tree_example}.

%as well as Figure \ref{fig:experiment_c_mermaid} in the appendix.

\subsection{Symbol Grounding: Words as Classifiers}
We use the \textit{Words-as-Classifiers} (WAC) model \cite{Kennington2015-ia} in the Symbol Grounding phase of our pipeline. WAC grounds an unlabeled category represented by its own classifier trained on positive and negative examples of sensory data from the region (negative samples being randomly sampled from other regions). WAC is well suited to our task, where unlabeled category has at most 30 observations, as it has been shown to perform well on small amounts of training data. %This is precisely the kind of use case that DINOv2 was designed for.

\section{Experiments}

In this section, we explain our experiments that build on prior work by incremental extensions. We first establish that Explauto can work with our framework and robot using YOLO+CLIP, then extend the curiosity model to make use of high-dimensional, visual vectors then use category learning (i.e., region splitting as explained above) using cosine similarity and the YOLO+CLIP then, SAM+DINOv2 pipeline. We first explain the procedure and metrics we use for each experiment (as they are the same for all three), then we explain the three experiments, then we compare the results.

\begin{figure}[h]
\centering
\includegraphics[scale=0.12]{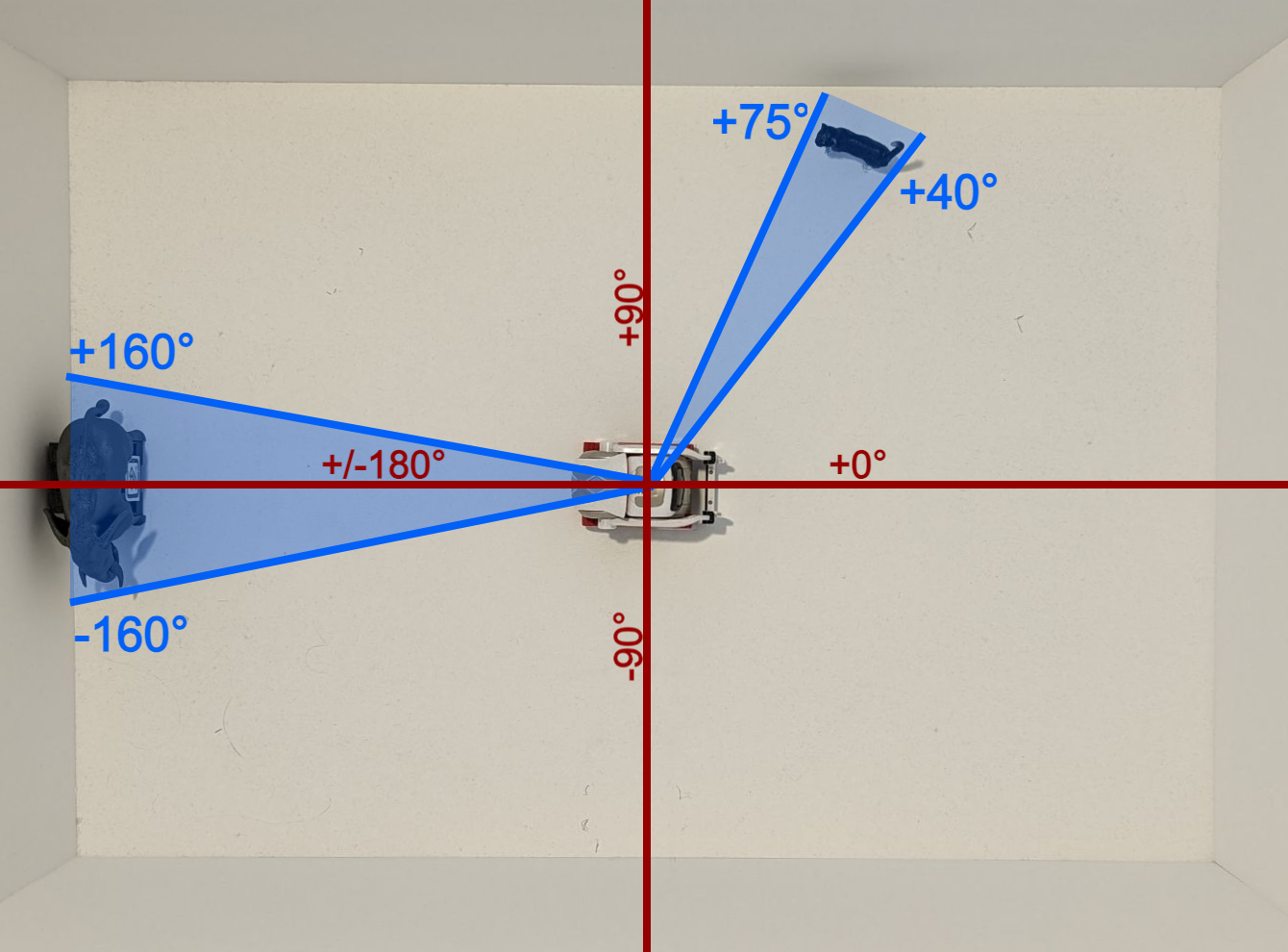}
\caption{The Experiment space used for our three experiments. The objects were located at approximately +55 degrees and and +/-180 degrees. The robot is initialized at at the center of the space facing 0 degrees.}
\label{fig:experiment_config}
\end{figure}

\paragraph{Procedure} All experiments took place in a constrained environment with a limited selection of objects; we use a toy cat and an elephant (see Figure \ref{fig:experiment_config}).\footnote{Our choice of using only two objects in a limited space for our experiments is important: we systematically establish that each of our extensions to prior work show improvement before we move onto the next experiment. Scaling our pipeline to work in noisier environments with a greater degree of movements and sensorimotor input is planned for the near future.} To counteract the gradual aggregate of positional error due to a 2 degree tolerance in Cozmo's rotation functionality, the robot re-centers itself after every fifth action.\footnote{On occasion, it is necessary to manually adjust a poor re-centering job which can be done during a brief pause before execution continues.} In order to support this automated re-centering, the elephant object has a symbol on the front which signifies to Cozmo that it can be used for docking. While the symbol has an impact on the CLIP or DINOv2 vector produced, we do not expect it to have a detrimental impact to the category development process. 
%Though, we have yet to see how this might effect the comparison of the resulting category with that of another elephant which does not have the symbol. 
% \todo{I'm not married to the last sentence but might be worth mentioning}

The experiment space measures 762x508 millimeters and has white floor and walls to reduce background complexity. This space allows for the robot to rotate up to $+/-$180 degrees and travel in a line up to $+/-$80mm. The objects can be placed anywhere in the experiment space, though it is beneficial for the object which doubles as a re-centering agent to be along the X or Y axis. While object locations remained static throughout the experiments outlined in this paper, we observed consistent category development no matter where objects were placed in the space.

% If Cozmo is near enough to an object such that it fills the entirety of the camera view, we found that the bounding boxes produced by YOLO were too inconsistent resulting in undesirable noise.\footnote{This noise might present as capturing a small section of shadow during one execution and not being able to identify anything the next} In an effort to reduce this noise, we were mindful to place the objects at a sufficient distance from the center which would limit how near to the object Cozmo could be when traveling the maximum linear distance of 80mm.

Each experiment consisted of 300 Action/Perception turns which lasted approximately three hours. After each action/perception turn, Cozmo returned to center before executing the next turn. For prediction, the Sensorimotor Model LWLR retrieved neighbors from the KD Tree. The Interest Model parameters had a max region size of 30, sampling epsilon (likelihood of random sampling) of 0.1, and Learning Progress Window Size (lookback window for calculating learning progress) of 10.

\begin{figure*}[h]
\centering
\begin{minipage}[c]{0.45\linewidth}
\centering
\includegraphics[scale=0.2]{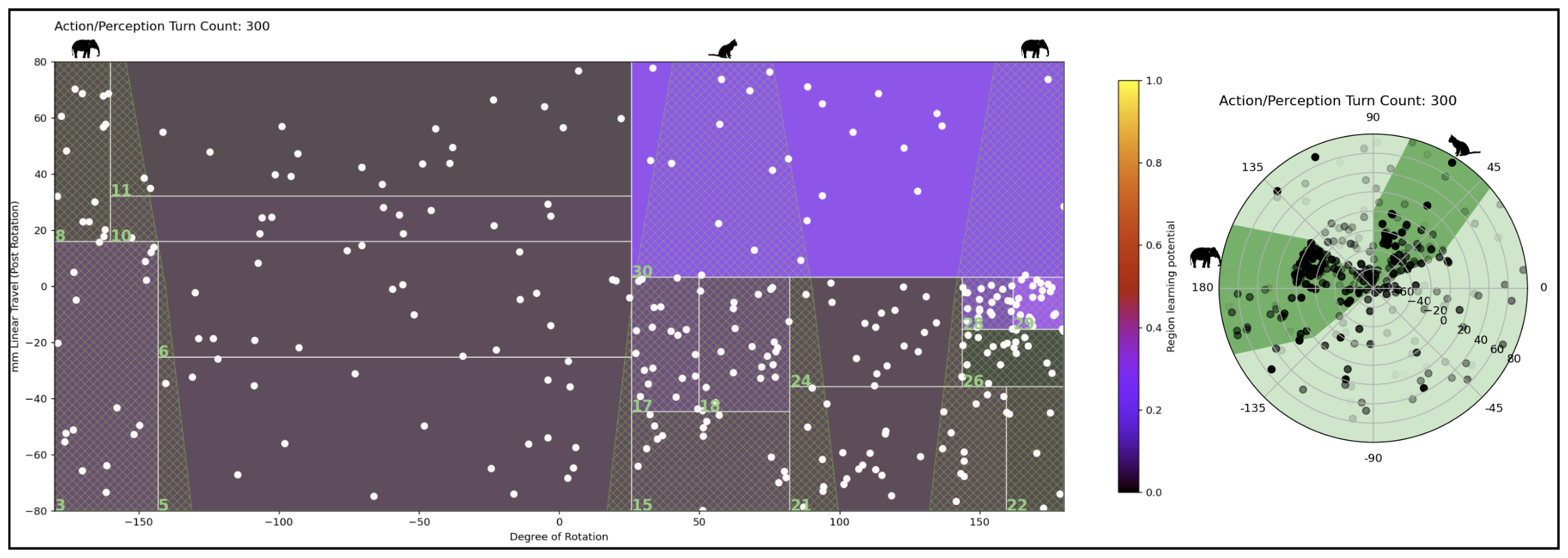}
\caption{Results from an execution of Experiment 2. }
\label{fig:experiment_c_plots}
\end{minipage}
\hspace{0.3cm}
\begin{minipage}[c]{0.45\linewidth}
\centering
\includegraphics[scale=0.2]{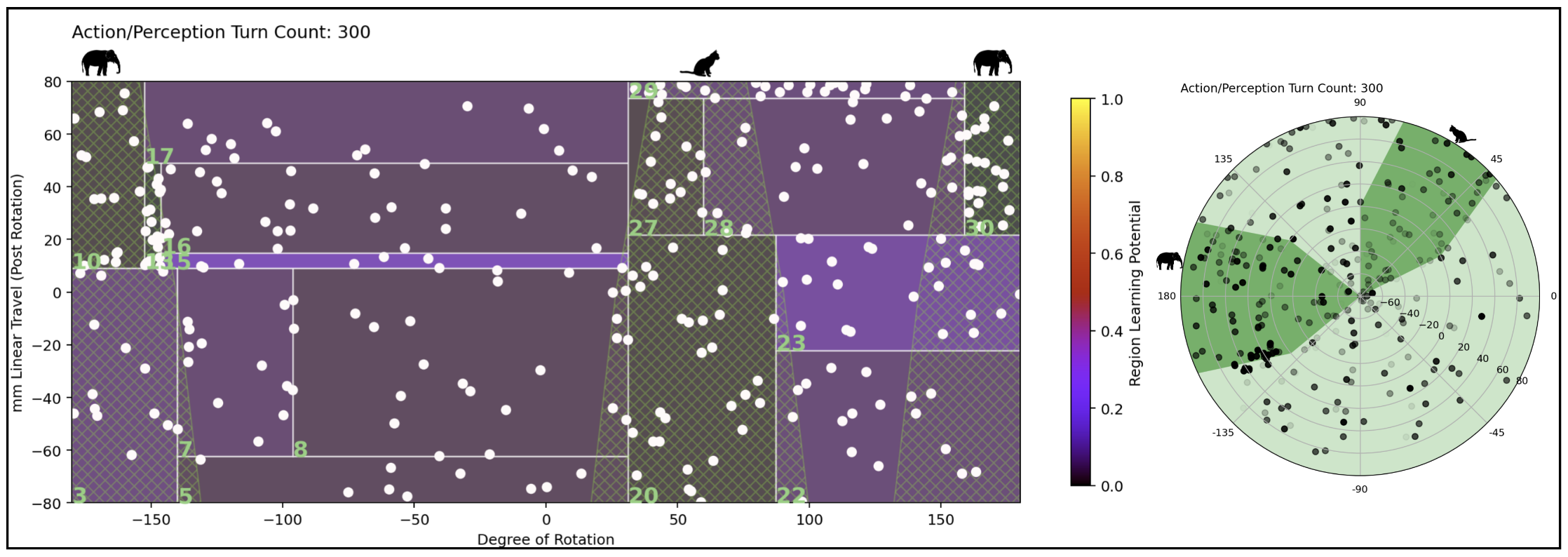}
\caption{Results from an execution of Experiment 3. }
\label{fig:experiment_d_plots}
\end{minipage}
\end{figure*}

\paragraph{Metrics} 
Throughout our experiments, we observe where in the Interest Space each action is sampled from, how each action impacts the learning progress for a region, and the value along which the axis is split during region splitting. For every action taken there is a target (predicted) and reached (actual) sensory outcome, as defined in Explauto. These sensory outcomes are used in calculating `competence' which is used in calculating the region's learning progress which is used in determining the next action to sample. As each experiment progresses, this cycle results in the development of sub-regions throughout the space which we refer to as unlabeled categories. For each experiment, we capture these values in figures to better understand the category boundaries.

The grid plot (Figure \ref{fig:experiment_a_plots}, left) provides a view into the region splitting and action sampling mechanisms. Each point represents an action taken, with the degree of rotation along the x axis and mm linear distance traveled (post rotation) along the y axis. In the grid plot to the left the sub-regions (unlabeled categories) have a much closer alignment to the objects in the space. The polar plot to the right shows recent motor goals are largely concentrated in areas where an object will be perceived. There are three shaded areas marked on the plot which roughly denote actions which will result in an object being in Cozmo's FOV. The boundary of each region is marked by a white line with a number reflecting the unlabeled category in the bottom right. As the experiment progresses, the color of each region dynamically updates to reflect the current learning progress as seen in Figure \ref{fig:learning_potential_progression}. The polar plot (Figure \ref{fig:experiment_b_plots}, right) provides a view into where in the Interest Space more recent actions have been sampled. The darker the point denotes a more recent action, which allows us to observe if and when the exploration begins to narrow in on objects in the space; shaded areas marked to signify the objects. Comparisons can easily be made visually using these plots.

In our final evaluation, we compare the categories produced by the pipelines that use YOLO+CLIP and SAM+DINOv2 by comparing the accuracy of unlabeled WAC classifiers to identify objects within a category. 

\paragraph{Baseline Setup: Extending Explauto with Cozmo}

We first establish that Explauto can be integrated into the Retico pipeline using Cozmo as the physical robot (i.e., the ability to physically move as well as visually observe its environment---an extension from prior work) while changing as little as possible in the Explauto framework. As such, we used YOLO only in determining the binary presence of an object. This aligned most closely with the original Playground Experiment for Explauto. For a given motor action of \textit{rotation} or \textit{linear travel}, the sensory output is 1 when an object was detected and -1 otherwise. The results of this experiment are shown in Figures \ref{fig:learning_potential_progression} and \ref{fig:experiment_a_plots}. The figures show that the novel sensorimotor space and Cozmo robot are usable with Explauto, setting the stage for our experiments.

\subsection{Experiment 1: Sensing a high-dimensional, visual sensory space with YOLO+CLIP}

In this experiment, we go beyond prior work by increasing the richness of the sensory space by incorporating the full CLIP+ feature vector as the sensory output for a given motor action (prior work only was concerned with binary presence of an object, not the visual representation of an object). However, prior work only used very small vectors (smaller than size 10), whereas here we are working with much bigger vectors that can vary in size depending on the model (i.e., CLIP produces vectors of size 512, DINOv2 of size 384). Therefore, to support the higher dimensional vectors and variability of their sizes, we alter the Competence calculation from Euclidean Distance to a measure of Cosine Distance formalized as \begin{equation}
1 - \cos ({\bf t},{\bf e})= {{\bf t} {\bf e} \over \|{\bf t}\| \|{\bf e}\|} = \frac{ \sum_{i=1}^{n}{{\bf t}_i{\bf e}_i} }{ \sqrt{\sum_{i=1}^{n}{({\bf t}_i)^2}} \sqrt{\sum_{i=1}^{n}{({\bf e}_i)^2}} }
\end{equation} 

\noindent
We bound this distance between (0,1) by $(1 - e^{2 * -cos\_dist})$.

The results of this experiment can be viewed in Figure \ref{fig:experiment_b_plots}. The grid plot shows that the sub-regions align with the objects in space with fewer sub-regions in non-object areas than the Baseline. Moreover, the polar plot shows that motor goals coalesce around objects rather than whitespace.

\subsection{Experiment 2: Region Splitting using Cosine Similarity with YOLO+CLIP}

In this experiment, we build on Experiment 1 but instead of splitting by the value which maximizes the difference of learning potential in the two sub-regions, we explore a new region splitting mechanism. This new region splitting mechanism divides the regions such that the \textit{variance of the cosine similarity} is minimal on each side, encouraging region splitting along visual object boundaries. Importantly, this experiment results in the robot learning specific categories of object features and location from the CLIP+ vector---the autonomous bottom-up category discovery that can later be symbolically grounded into.

The results of this experiment can be viewed in Figure \ref{fig:experiment_c_plots}. In this plot, we observe much closer alignment to the objects in the space, and the motor goals are highly concentrated in areas where an object will be perceived.

\subsection{Experiment 3: Region Splitting using Cosine Similarity with SAM+DINOv2}

\begin{figure*}[h]
\centering
\includegraphics[scale=0.3]{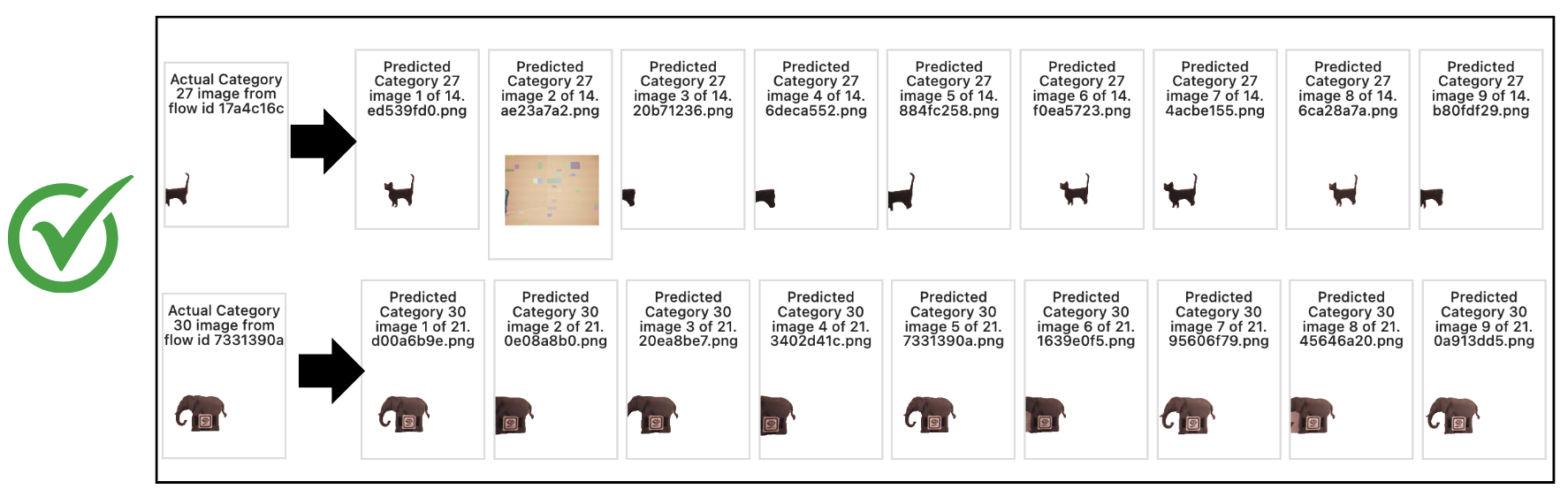}
\caption{Given a sensory exemplar from an unlabeled category, WAC is able to predict the correct category. On the left is the photo used to produce the sensory exemplar, on the right are images from the region of the unlabeled category. The upper correctly predicts Category 27 and the bottom Category 30.}
\label{fig:wac_exp_d_correct_predictions}
\end{figure*}

\begin{figure*}[h]
\centering
\includegraphics[scale=0.3]{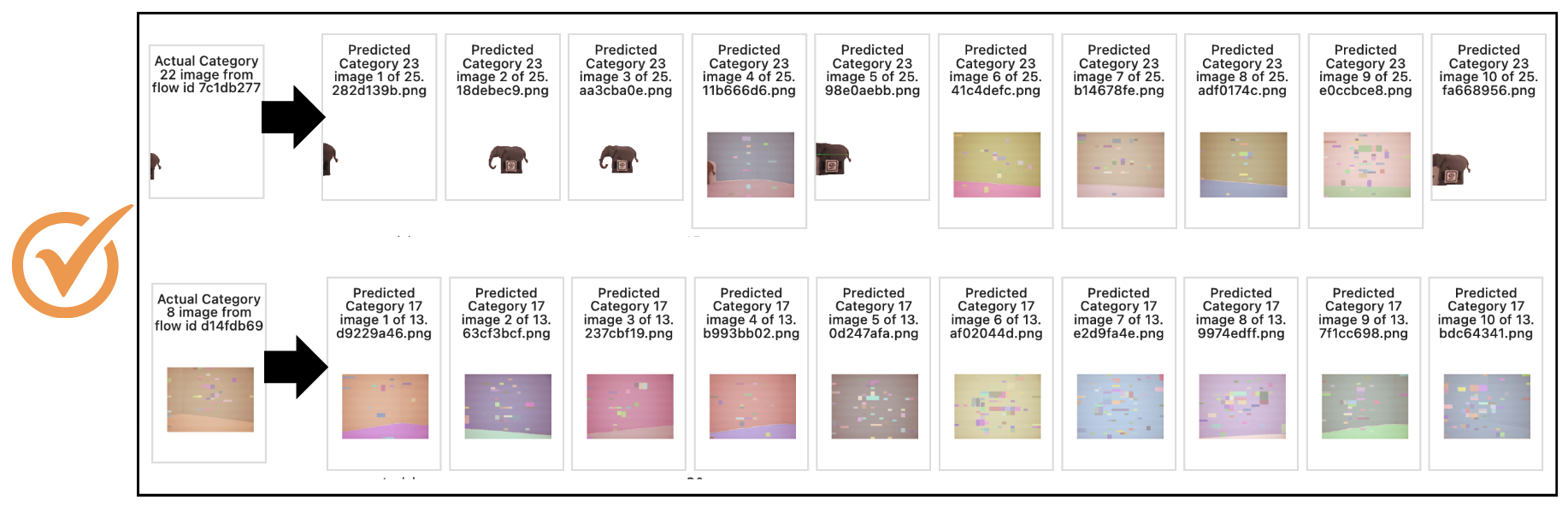}
\caption{Given a sensory exemplar from an unlabeled category, WAC is not able to predict the correct category but the images from the region of the predicted unlabeled category are nearly identical to the sensory exemplar and the categories shown in the associated grid plot cover a similar range of objects (seen in \ref{fig:experiment_d_plots}. We consider this to be correct on a technicality and liken it to human error. The upper row predicted Category 23 for an exemplar from Category 22 and the bottom predicted Category 17 for an exemplar from 8.}
\label{fig:wac_exp_d_correctish_predictions}
\end{figure*}

\begin{figure*}[h]
\centering
\begin{minipage}[c]{0.6\linewidth}
\centering
\includegraphics[scale=0.3]{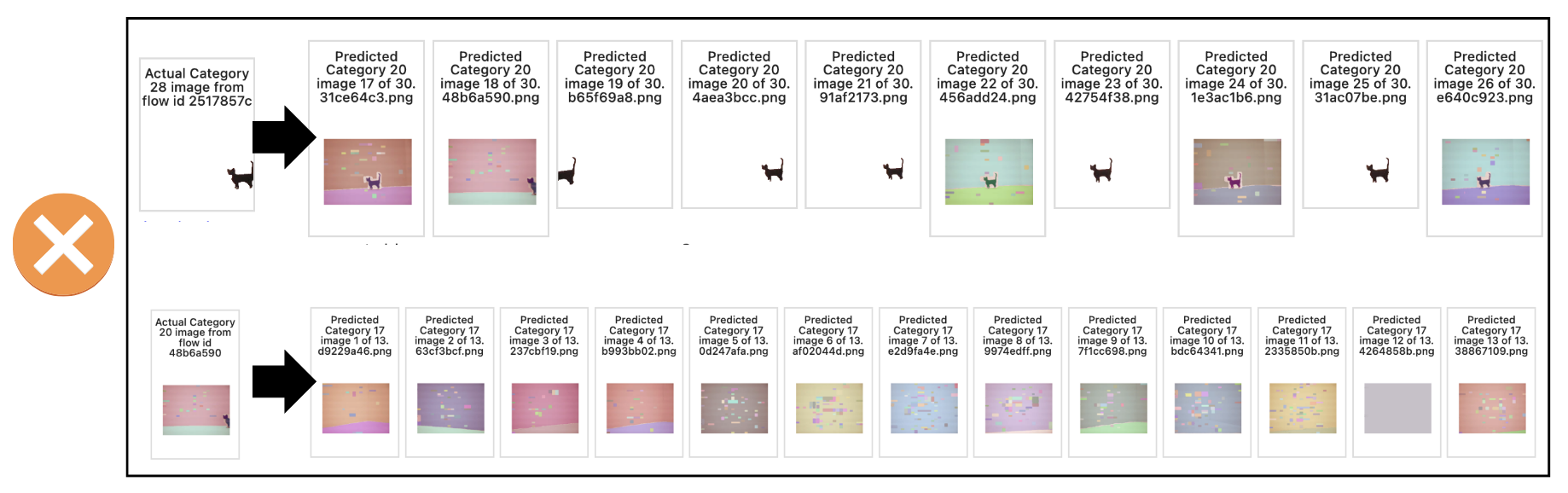}
\caption{Given a sensory exemplar from an unlabeled category, WAC is not able to predict the correct category. The first row is an edge case where the predicted unlabeled category 20, for an exemplar from 28, consists of both cat and ``whitespace" observations (``whitespace" is shown by images with SAM segment overlays) and the region seen in the associated grid plot does not cover a similar range of objects as 28 (seen in \ref{fig:experiment_d_plots}). The bottom row, which predicted Category 17 for exemplar 20, captures the edge case where there was a limitation in SAM and a cat in view was not successfully segmented and/or isolated resulting in a false ``whitespace" observation. sensory exemplar of $\mathbf{[-1]*384}$.}
\label{fig:wac_exp_d_incorrect_predictions}
\end{minipage}
\hspace{0.5cm}
\begin{minipage}[c]{0.25\linewidth}
\centering
\includegraphics[scale=0.13]{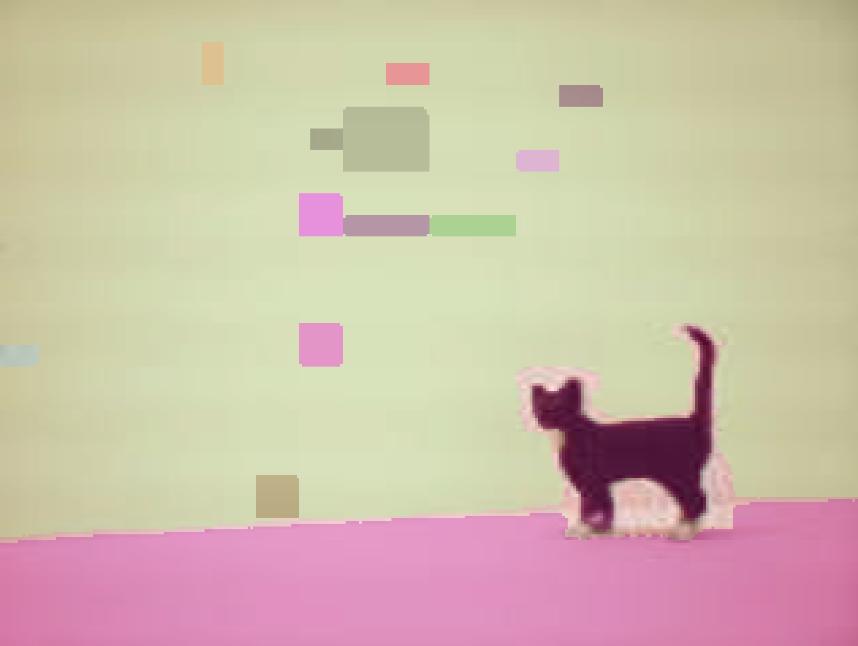}
\caption{SAM having successfully segmented the wall and floor, but missing the cat. There is also a fair amount of  segments which don't appear meaningful though may be due to light bounce or noise in the image.}
\label{fig:sam_segmenting_missed_cat}
\end{minipage}
\end{figure*}

In this experiment we build on Experiment 1 and 2 by replacing the object detection and representation from YOLO+CLIP to SAM+DINOv2 to move closer to our goal of learning in a noisier environment. Because SAM identifies many segments for a given frame, it was necessary to implement a method to reduce the noise which would occur if non-object segments were isolated and passed down the pipeline despite an object being in view. Such noise results in inconsistent exemplars for a region and have significant negative downstream impact to the ability to correctly assess novelty. To prevent this noise we only use an isolated segment if the average pixel color is above 191.25. This threshold value was determined by observing the average pixel value for segments consisting of only the whitespace background of our environment which does have a reddish hue from the Cozmo camera. Additionally, as shown in \ref{fig:sam_obj} the extracted image is no longer a tight crop around the object. 

% Because we are working with masks instead of bounding boxes, we lack the information necessary for the supplementary positional information we had added in Experiment 1 and no longer include it. Moving to DINO v2 results in the feature vector decreasing from size 512 to 384. Despite these changes the resulting categories appears to be of similar precision and coverage to that of Experiment 2. 

The results of this experiment can be viewed in Figure \ref{fig:experiment_d_plots}. The grid-plot shows a similar precision in constructing category boundaries as seen in Experiment 2. The polar plot, however, deviates from the other experiments in that actions were far less clustered in the observation space. Instead, actions were sampled more broadly throughout the observation space, even late into the execution, with a slightly heavier concentration in areas where an object will be perceived.

\subsection{Discussion}
The grid plots from the Baseline and Experiment 1 both show a very loose relationship between region and objects in the space, as expected, but they both establish that the framework with the added functionality works in a similar way as prior work. Notably, however, the region splitting implementation introduced in Experiment 2 results in region development with far more precision along object boundaries. Improving this relationship between regions and objects in space allows us to treat the regions as category boundaries and consequently leverage the sensory data associated with each region as a representation of that category. The grid plot in Experiment 3 shows similar precision in category boundaries as those seen in Experiment 2. 

The polar plots from each experiment show the points from more recent observations as a darker shade of black. From this, we can determine that Experiment 2 increasingly samples motor goals which result in the observation of an object. This is promising behavior as we are looking to capture rich sensory representations of the unlabeled categories and there is little to no benefit in collecting more non-object data points than is necessary to determine the object boundary. It is interesting to observe that Experiment 3 does not appear to have the same level of honing in on actions resulting in object observation.

\subsection{Final Evaluation: Symbol Grounding into Categories}

We evaluate the quality of the unlabeled categories from Experiments 2 and 3 using WAC. The goal is for WAC to predict the category name given a sensory exemplar from the Interest Space. Each experiment resulted in 16 categories (making a random baseline of 6.3\%); each category had a range of 8 and 30 samples, average of 18.75. We randomly sampled 80\% from each for training and 20\% for testing. To train WAC, we train a binary logistic regression classifier for each category using observations from a specific category as positive examples, and randomly sample observations from neighboring categories as negative samples. We report the number of times WAC was able to accurately predict a category both when using exemplars from Experiment 2 in training and test, and separately, Experiment 3. This step is crucial because if classifiers can distinguish between the categories, then that means they are ready to have a word label attached to them, completing the word-level symbol grounding for that category (though at this step, the process is not interactive; we leave that for future work).

Table~\ref{tab:wac2} shows the results using the categories from Experiment 2. There were 5 correct Category predictions which require no explanation. The rows marked SC denote scenarios where the actual and predicted categories were visually similar and we consider them to be ``Similar Categories". The rows marked YOLO would appear incorrect at a glance, but the point of failure is YOLO rather than Explauto.

\begin{table}[]
\centering
\scalebox{0.9}{
\begin{tabular}{@{}lll@{}}
Actual Category & Predicted Category & Manual Explanation \\
\rowcolor[HTML]{EFEFEF} 
5      & 24        & SC          \\
3      & 3         &             \\
\rowcolor[HTML]{EFEFEF} 
10     & 24        & SC          \\
15     & 18        & SC          \\
\rowcolor[HTML]{EFEFEF} 
11     & 24        & SC          \\
28     & 24        & YOLO        \\
\rowcolor[HTML]{EFEFEF} 
18     & 17        & SC          \\
6      & 24        & SC          \\
\rowcolor[HTML]{EFEFEF} 
30     & 30        &             \\
8      & 8         &             \\
\rowcolor[HTML]{EFEFEF} 
29     & 29        &             \\
21     & 24        & YOLO        \\
\rowcolor[HTML]{EFEFEF} 
26     & 24        & YOLO        \\
24     & 24        &             \\
\rowcolor[HTML]{EFEFEF} 
17     & 18        & SC          \\
22     & 29        & SC         
\end{tabular}
}
\caption{Results from WAC predictions trained on the Unlabeled Categories from YOLO+CLIP in \textbf{Experiment 2}.}
\label{tab:wac2}
\end{table}

Table~\ref{tab:wac3} shows the results using the categories from Experiment 3. There were 6 correct Category predictions which require no explanation. The rows marked SAM would appear incorrect at a glance but the point of failure is SAM rather than Explauto. The rows marked IPCF, we theorize, are incorrect due to ``in progress category formation". Categories 28 and 20 show a split along object/whitespace boundaries, category 28 due to natural object/whitespace boundary and category 20 due to SAM failing to segment objects at a distance.

As evidenced by the manual explanation labels, evaluation with WAC is not yet a fully automated process. This is further demonstrated by Figure \ref{fig:wac_exp_d_correctish_predictions} where a sensory exemplar from a particular category is predicted as being from a different category. Looking at the images of the exemplar vs those constituting the predicted category, they are nearly identical. Moreover, the grid plot from Figure \ref{fig:experiment_c_plots} shows that the two categories are visually similar to one another in the Interest Space, both overlapping where the similar features in the exploration space,  elephant (row 1) or whitespace (row 2). Because the predicted vs actual categories are so similar, so much so that it can be likened to human error, we consider this to be a technically correct prediction. Figure \ref{fig:wac_exp_d_incorrect_predictions} has examples of incorrect predictions,  row 1 due to underdeveloped categories and row 2 due to inconsistencies in SAM segmentation.

Taken together, these results show that both pipelines work to allow the robot to curiously explore its space and arrive at unsupervised object categories. In the YOLO+CLIP pipeline, the object categories tended to be cleaner (as expected), but the noisier and more theoretically pleasing SAM+DINOv2 pipeline worked well; we conjecture that it simply needs more exposure to the objects to learn more rigid categories. This also highlights the robustness of our pipeline; theoretically, any kind of object detection model and object representation model could be used because the cosine distance calculation is agnostic to the vector size from the representation model.

% \begin{figure*}[h]
% \includegraphics[scale=0.35]{wac_correct_predictions.png}
% \caption{Given a sensory exemplar from an unlabeled category, WAC is able to predict the correct category. On the left is the photo used to produce the sensory exemplar, on the right are images from the region of the unlabeled category. The upper correctly predicts Category 3 and the bottom Category 8.}
% \label{fig:wac_correct_predictions}
% \end{figure*}

% \begin{figure*}[h]
% \includegraphics[scale=0.35]{wac_correctish_predictions.png}
% \caption{Given a sensory exemplar from an unlabeled category, WAC is not able to predict the correct category but the images from the region of the unlabeled category are nearly identical to the sensory exemplar. We consider this to be correct on a technicality and liken it to human error.}
% \label{fig:wac_correctish_predictions}
% \end{figure*}

% \begin{figure*}[h]
% \includegraphics[scale=0.35]{wac_incorrect_predictions.png}
% \caption{Given a sensory exemplar from an unlabeled category, WAC is not able to predict the correct category. Both rows capture edge cases where YOLO did not detect the object in view anything which resulted in a whitespace sensory exemplar of $\mathbf{[-1]*519}$.}
% \label{fig:wac_incorrect_predictions}
% \end{figure*}

Figure \ref{fig:wac_exp_d_correct_predictions} captures examples where WAC was able to correctly predict the unlabeled category. The first row correctly predicts a category which appears to capture \textit{cat-ness} at a nearer proximity to Cozmo, as seen in the grid plot in Figure \ref{fig:experiment_d_plots}. The second row appears to capture \textit{elephant-ness} up close, also as seen in the grid plot.

\begin{table}[]
\centering
\scalebox{0.9}{
\begin{tabular}{@{}lll@{}}
Actual Category & Predicted Category & Manual Explanation \\
\rowcolor[HTML]{EFEFEF} 
8      & 17        &  SC         \\
30      & 30       &             \\
\rowcolor[HTML]{EFEFEF} 
28     & 20        &  IPCF       \\
3      & 3         &             \\
\rowcolor[HTML]{EFEFEF} 
20     & 17        &  SAM        \\
23     & 17        &  IPCF       \\
\rowcolor[HTML]{EFEFEF} 
13     & 13        &             \\
29     & 17        & IPCF        \\
\rowcolor[HTML]{EFEFEF} 
17     & 17        &             \\
10      & 10       &             \\
\rowcolor[HTML]{EFEFEF} 
16     & 17        & SC          \\
5      & 17        & SC          \\
\rowcolor[HTML]{EFEFEF} 
7      & 17        & SC          \\
27     & 27        &             \\
\rowcolor[HTML]{EFEFEF} 
22     & 23        &  SC          \\
15     & 17        &  SC        
\end{tabular}
}
\caption{Results from WAC predictions trained on the Unlabeled Categories from SAM+DINOv2 in \textbf{Experiment 3}.}
\label{tab:wac3}
\end{table}

\section{Limitations}

\paragraph{SAM}
In the current implementation of Experiment 3 we are able to iterate through the segments and isolate only those which captures the object in frame (as opposed to a segment for wall, floor, or miscellaneous noise). As is, this can not be scaled to a real-world scenario outside of the controlled experiment space because the average pixel threshold is meaningless outside of a known and consistent background. Working with unlabeled segments trained only using image data is applicable to our long-term goals, however it does come with its challenges. The foremost is that there are limited ways to ensure the same object or feature within the exploration space for a given motor action is consistently segmented, especially in an object rich space. Some of the complexities of SAM segmentation can be seen in Figure \ref{fig:sam_segmenting_missed_cat}.

\paragraph{Robot Movement Drift}
For the purposes of our experiments, we always re-centered Cozmo from a starting position so we could systematically compare our model's decision as to which action to take next. Cozmo unfortunately drifts when it turns (within a tolerance of 2 degrees), requiring us to use the symbol (see Figure~\ref{fig:sam_obj}) to re-orient Cozmo. This does not seem to interfere with our results, but ideally this would not be required (though it is far better than manually re-centering Cozmo after each step). This constraint is only applicable, however, to our experimental setup; a setting where Cozmo can roam freely in a space would not require strict re-centering.

% \paragraph{Robot Movement}
% For the purpose of our experiments, it is expected that that each action takes place from the initial pose of 0 degrees of rotation and 0 mm of linear travel so we can systematically compare each decision. As a result, if the robot fails to recenter itself after each action it is possible that the intended rotation of 40 degrees is actually 43 or even 50 degrees from 0. The turn in place functionality from the Cozmo SDK we use has tolerance clamped at a minimum of 2 degrees, and unfortunately, adjusting that is beyond our control. As a result, over the course of an execution it is possible for Cozmo to drift from center as the error in the rotation accumulates which results in incorrect predictions and false novelty in the system. The intermediate solution for this problem is to leverage built-in functionality for Cozmo to align itself to one of the provided cubes. Because we did not want the cube to interfere with the object space, we elected to embed the cube into one of the objects we use for our experiments. In he long-term, we would like to remove the need for recentering altogether, preferring to iterate in a continuous action space where each movement is an extension of the last. \footnote{Cozmo docs outlining the clamping on turn\_in\_place() tolerance can be found here: \url{https://data.bit-bots.de/cozmo_sdk_doc/cozmosdk.anki.com/docs/generated/cozmo.robot.html\#cozmo.robot.TurnInPlace}. At the time of writing the DDL servers are no longer up, so it is not possible to access the official docs.}

\section{Conclusion and Future Work}
Through visual analysis of the output plots as well as our final evaluation with WAC, our experiments illustrate the robot is able to autonomously explore the space, constructing unsupervised, unlabeled categories in a bottom-up progression. It is important to emphasize here that these categories have never had a label manually assigned; rather, they develop bottom-up based on the robot's Sensorimotor experience and we simply used WAC to see if it could distinguish between the (still unlabeled) categories. Categorization is a crucial step of the language learning process and, more generally cognition because ``to cognize is to categorize" \cite{Harnad2017-jo}. 

In future work, we will address the limitations explained above. We will also include scenes with more objects, more complex objects, and a larger space for Cozmo to explore. We will also include additional sensorimotor information as well as increasing the actions that Cozmo can take. 

% Upon addressing the limitations, we will continue exploring the unlabeled categories and in this regard we have many ideas. 

% \begin{itemize}
%     \item Iterate on how we can use WAC as an evaluation metric
%     \item Walk the Interest Model Tree and consolidate categories
%     \item Label the categories interactively with human "teacher"
%     \item Infer relationships
%     between categories, both labeled and unlabeled. 
%     \item Use weights from the trained WAC model 

% \end{itemize}

%% Use plainnat to work nicely with natbib. 

\bibliographystyle{plainnat}
\balance
\bibliography{sample-base,paperpile}

\end{document}